# New Conditions for Philosophers to Catch the Wave of Citizen Deliberation in the Age of Artificial Intelligence

**Bernard Reber**

*National Centre for Scientific Research and*
*Centre de recherches politiques de Sciences Po*

**Abstract**: Powerful technologies labelled "AI" — without sufficient epistemic caution — are already reshaping political and private life, bringing both new dangers and new opportunities for citizen participation. These range from electoral and legislative engagement to the most ambitious form: political co-creation through citizens' assemblies.
Large Language Models (LLMs) could support such processes through moderation, translation, facilitation, summarization and writing assistance. But this potential remains largely unrealized.
The *Democratic Commons project* takes a fundamentally interdisciplinary approach — from philosophy to computer science — to evaluate LLMs against five proposed democratic principles. At its core, the project is driven by the question of political bias: under what conditions can LLMs be used democratically within forms of citizen participation that are themselves still largely experimental?
Addressing these socio-technical questions requires grounding in political theory and, more broadly, in philosophy — disciplines that provide the normative frameworks without which the democratic evaluation of AI systems cannot be meaningfully conducted.

The objective of my title is to respond to the following question: what is the purpose of philosophy in our age? This question is particularly relevant in the context of a technology that claims to touch on intelligence through the way it is named. Without too many epistemic precautions, it is called artificial intelligence. The question of usefulness is not trivial. For instance, this is the title of the latest book (2) by Philip Kitcher: *What's the Use of Philosophy*?[1] In considering Tomonobu Imamichi's insights regarding philosophy and eco-ethics in the context of technological conjuncture, the question arises: how would he respond to the challenges posed by Artificial Intelligence (AI) in the present era, given his initial reservations about the telephone, an old technology?

In light of this, I will be presenting the general conceptual framework of Democratic Commons (DC)[2], a project for which I serve as one of the Research Director. DC is at the intersection of AI and democracy, and more precisely participatory democracy. It is ambitious in nature and interdisciplinary in scope, involving more than thirty researchers, all post-doctoral fellows from Sciences Po and Sorbonne University, in conjunction with associates from Yale and Harvard and many AI players. About twenty practitioners are also involved in developing platforms for new forms of citizen participation.

1. Philip Kitcher, *What's the Use of Philosophy?*
2. See: https://about.make.org/democratic-commons/landing-page#founding-members

The final point in the title of my article refers to metaphor used by an OECD report[3] that speaks of a *wave* for participatory and deliberative experiments, bringing together randomly selected citizens and experts to inspire new technologies assessment or public policy. It is acknowledged that this wave is predominantly European, mainly concerning citizens' assemblies on climate. This phenomenon stands in contrast to the lack of interest shown by high-level decision-makers in the United States, for example, during both the Trump and Biden administrations, in the proposals put forth by informed citizens through deliberative processes in pluralistic assemblies. This perspective is further substantiated by the findings of James Fishkin of Stanford University,[4] with whom we are in collaboration. These polls, while valuable in their own right, remain confidential and serve merely as an interactive survey of citizens' thoughts, offering no opportunity for the co-construction of proposals, which are, as we have previously argued,[5] much more complex and difficult to achieve. In the Democratic Commons project, we will explore many other opportunities for participation throughout democratic life, thanks to digital platforms designed for civic engagement.

Finally, I am skeptical about public intellectuals, a role frequently adopted by philosophers, who purport to discuss a wide range of subjects. Scientists who respect themselves would not engage in such behavior, nor would they stray into areas outside their area of expertise. The tendency to engage in high-minded discourse, as (3) exemplified by the likes of Sartre, is to be avoided, as it can prove to be a source of profound disquiet. Instead, what is required is a contribution that is both relevant and within a system, both academic and extra-academic.

Conversely, my intention is not to defend a philosophy solely for philosophers and within this multifaceted discipline alone. I believe that philosophy, particularly political philosophy, can benefit from analyses of practical experiences. I have done this, for example, by discussing both philosophical points and duly analyzed case studies.[6]

We could say, with Albert Einstein, that science without experience is incomplete. Without losing sight of the specific requirements of philosophy, particularly normative ones, we can thus be guided in the analysis of practical cases and, if necessary, revisit these frameworks after this analysis.

In the cases presented here, political philosophy serves above all to create ex ante the conditions for these participatory platforms to meet democratic requirements and respect democratic principles. The first step is therefore to produce and defend democratic principles and to deduce indicators and questions from them. These should serve as guidelines for other members of the project working in various disciplines (sociology of practices, ethnography, computer science). This requires risk-taking and interdisciplinary interpretations and encounters limitations that we will revisit in our conclusion.

More continuous citizen participation in the various stages of democratic life is already an issue in itself (part 1.). Its evaluation remains in its infancy.[7] If, in addition, access to this participation is provided by platforms where participants express themselves very briefly (part 2.), the conditions of their participation pose an additional problem. Admittedly, they are very similar to the habits of so-called social networks in their rather limited and non-interactive nature. A final problem is the emergence of large language models (LLMs), which are now used as resources on these platforms. There are more and more platforms and their LLM-enabled experiences and research in this interface between democracy and AI. However, very few of them have a clear idea of the democratic principles that should guide such tools.

---

3. OECD, *Innovative Citizen Participation and New Democratic Institutions: Catching the Deliberative Wave*.
4. Alice Siu and Bernard Reber, "Climate Conversations: A Comparative Study of Citizen Deliberations in FR and US," https://francestanford.stanford.edu/projects/climate-conversations-comparative-study-citizen-deliberations-fr-and-us.
5. Dimitri Courant and Reber, eds. *Deliberative Democracy and Ecological Transition:. The French Citizens' Convention for Climate*.
6. Reber, *Precautionary Principle, Pluralism, Deliberation: Science and Ethics*, and *Responsible Deliberation, between Conversation and Consideration*.
7. Reber, *Responsible Deliberation*. Courant and Reber, *Deliberative Democracy and Ecological Transition*.

The philosophy of language and political philosophy therefore have an important role to play in criticizing the limitations of most platforms and AI-mediated experiences that claim to be democratic without democratic principles (part 3.), but also in attempting to limit certain biases and offer better solutions.

What we are dealing with is more of an engineers'democracy, which offers improved surveys rather than deliberative tools. The DC project, on the other hand, works on the basis of purely literary technologies—translating, synthesizing, helping to formulate—and participation—moderating, facilitating. (4)

## 1. AI and Democracy: Between Perils and Promises

A range of technologies aimed at achieving artificial intelligence *effects*, backed up by the knowledge that makes them possible, is emerging in all fields. For the sake of readability, we will denote as AI this ensemble of technologies and enabling knowledge that is emerging across seemingly all fields. Politics is no exception. In this field, AI affects practices, particularly linguistic and discursive ones, as well as actors, institutions, and choices. Like any technology, AI entails new risks and threats, but it also offers unprecedented opportunities and potential advances. However, if we take a closer look, it is far more enigmatic[8] than its fascinating and somewhat racy name would suggest. We now know much better how to evaluate a translation between two languages, but we are far less advanced with the evaluation of a good synthesis of political debate that would reflect all opinions and a fortiori would help lead to democratic agreement. Linguistic norms, such as grammar and syntax, or those that take into account the contexts of enunciation, are not the norms of philosophy or political theory  that are implicit in all social sciences that study democratic life. Current controversies over the moderation of very large platforms, or the abolition of the moderation in the name of a very poor or even cynical interpretations of freedom of expression, indicate that the road to moderation in its full sense (organization of debate, equal consideration of proposals, search for justifications), which is essential to democratic deliberation, has yet to be realized. Furthermore, digital networks increasingly have the capabilities and power to disrupt society,[9] which is even more apparent as they incorporate AI into their functioning. Older technological means that made these digital networks/social networks/platforms possible have already disrupted the elaboration of knowledge and know-how, through impacts on data such as space, time, social interactions, and actor networks, and even knowledge from all the humanities and social sciences.[10] The very rapid changes in AI have far-reaching consequences, as they affect certain cognitive abilities by modifying the way individuals perceive, process, and integrate information, and therefore their ability to make political choices and participate in and guide collective action.

Democracy is a very specific way of making collective choices. It requires respect for very general values and normative principles; this respect is guaranteed by laws, institutions, rules of behavior, and an ethos that is more or less shared by the citizens who inherit them. While these laws reflect and interpret these principles in different ways, depending on national histories and their constant evolution, they nonetheless provide a framework for assessing gaps between countries in relation to them. They also enable us to situate the evolution of countries in terms of democracy. For (5) example, the ambitious V-DEM (Varieties of Democracy) project, inspired by work in political theory, documents the state of the world's democracies. We have taken this major contribution into account when comparing the state and evolution of democracy in all countries around the world. However, we will need to adapt it to these new uses.

---

8. Daniel Andler, *Intelligence artificielle, intelligence humaine: la double énigme*.
9. Yochai Benkler, Robert Faris, and Hal Roberts, *Network Propaganda: Manipulation, Disinformation, and Radicalization in American Politics*.
10. Reber and Claire Brossaud, eds., *Digital Cognitive Technologies: Epistemology and Knowledge Economy*.

AI must therefore be governed by democratic standards and respect certain values. This is already the case with the evaluation processes for any research project using AI, such as those funded by the European Commission.[11] Such projects must respond to a concern for ethical compliance, explaining what they will do and with what AI, but above all how they will ethically treat humans as well as data, guarding against possible misuse of the latter, both during and after the project. Ethical questioning concerns the very way in which algorithms are designed (principle of transparency), the knowledge of LLMs from their training data to the uses made of them by users and even the misuses made of them.

Similar concerns are guaranteed legally and more recently in all areas with the European Digital Act (2024) and more specifically the EU AI Act (2024), which came into force on February 2, 2025. These ethical and legal guarantees are therefore prerequisites for our project. They are necessary, but not yet sufficient or specific, to meet democratic requirements.

One of our most precious collective assets is democracy. It hasn't always existed, remains fragile, and is sometimes threatened without these technologies being at fault.[12] Added to this is the abundance of participation opportunities for citizens. These are provided by decision-making powers at the highest level (e.g., French Conventions; Conference on the Future of Europe) and by a host of initiatives (e.g., MetaGov) However, what may appear to be a democratic advance has to be evaluated and properly ordered if these forms of participation are to be democratic and relevant, all things considered. New technologies have also increased these possibilities. However, digital networks, all too entitled to the label "social," have increased these risks,[13] influencing the very way politicians talk to each other and the pace at which information flows. The arrival of LLMs is reconfiguring these issues, for better or worse. (6).

## 2. Thirteen New Democratic Uses

Thanks to practitioners from Make.org, who develop and moderate various civic platforms, we have been able to imagine new possibilities for participation in the project. I will briefly present them here with a short description, the hoped-for benefits of such participation, and, above all, how AI could be used as an assistant.

The DC project focuses on intermediation processes between one or more public institutions and citizens. Specifically, it concentrates on the following processes:

---

11. European Commission, *Ethics Guidelines for Trustworthy Artificial Intelligence*, https://digital-strategy.ec.europa.eu/en/library/ethics-guidelines-trustworthy-ai. Ibo Van de Poel, "Embedding Values in Artificial Intelligence (AI) Systems," *Minds and Machines* 30, no. 3.
12. Pierre Rosanvallon, *La contre-démocratie: La politique à l'âge de la defiance* and *Le bon gouvernement*. Marcel Gauchet, *La démocratie contre elle-même*. James Fishkin, *Can Deliberation Cure the Ills of Democracy?*
13. Shoshana Zuboff, *The Age of Surveillance Capitalism: The Fight for a Human Future at the New Frontier of Power.*

- Local and national electoral phases—participation in and monitoring of campaigns, organization and monitoring of public debates, citizen information, accessibility of candidates programs and proposals, etc.
- Local and national governance phases—participation in public policies at the local or the national level, participation in the development and implementation of these policies, ensuring transparency of the decision-making processes, evaluation of public policies, etc.
- National legislative process—participation in the drafting and monitoring of laws, improved transparency and support to parliamentary debates monitoring, improved understanding of laws, etc.
- Specific participatory processes—participation in a citizens' convention or a participatory budget.

## 2.1 Electoral Phases (Local and National)

### *2.1.1 Participation in the Citizen Agenda to Inform Programs and Debates*

- Description—enabling all citizens to articulate, ahead of elections, the key themes and priorities that can inspire political debates and platforms.
- Benefits for institutions and processes—reconnecting political debates and platforms with citizens' concerns, getting better citizen engagement, improving voter participation, and reducing the impact of fake news.
- AI use example—AI solution that allows citizens to identify their top ten most agreed-upon priorities ahead of a presidential election.

### *2.1.2 Understanding of and Participation in Candidates' Programs*

- Description—ensuring all citizens can understand and react to the platforms of candidates and political parties, compare them, and track their evolution.
- Benefits for institutions and processes—improved accessibility to parties' platforms, increased importance and comparisons of platforms in the electoral campaign, fostering greater engagement of citizens, better participation in the election, reduced impact of fake news. (7).
- AI use example—AI tool providing citizens with a summary of the official platforms of all the candidates and an analysis of their evolution, enabling the citizens to react in real time.

### *2.1.3 Follow-up and Reactions to Candidate and Party Debates*

- Description—enable all citizens to follow, understand, and react to the reality of candidates' words.
- Benefits for institutions and processes—improved citizen access to the candidates' statements and debates, promoting clear and accessible political discourse, fostering greater citizen engagement, improved voter participation, reduced impact of fake news, and improved media behavior of candidates.
- AI use example—an AI tool that provides citizens with an objective and comprehensive overview of all the statements made by the candidates (or parties) during a presidential election campaign, enabling real-time feedback to enrich debates.

## 2.2 Phases of Governance from Local to National Level

### *2.2.1 Monitoring the Implementation of Electoral Platforms*

- Description—enabling all citizens to track the implementation of public policies promised by the elected candidate in their platform.

- Benefits for institutions and processes—enhancing citizen engagement in public action, increasing citizen ownership of public policies, strengthening the executive's capacity to act, reducing impact of fake news.
- AI use example—AI tool providing a summary of executive and legislative actions throughout a mandate, compared to the elected candidate's platform.

*2.2.2 Monitoring Public Debates*

- Description—enabling all citizens to follow and understand public debate, in particular the statements made by elected officials.
- Benefits for institutions and processes—improved citizen engagement and trust in public action, improved ability of the executive to act, fostering better media behavior among elected officials, and reducing impact of fake news.
- AI use example—an AI tool delivering citizens a synthesis based on statements made by members of a municipal council in a large city.

*2.2.3 Participation in the Prioritization, Development, and Implementation of Public Policies*

- Description—enabling all citizens to participate in all stages of public action, from inception to co-construction. (8)
- Benefits for institutions and processes—enhanced citizen engagement in public action, greater citizen ownership of public policies, improved public policies, improved ability of the executive to act.
- AI application example—an AI platform that facilitates cycles of consultations and deliberations, allowing citizens to co-create transportation policies for a given locality.

*2.2.4 Understanding and Interacting with Executive Decision-making Processes*

- Description—empowering all citizens to understand and engage with executive decision-making processes.
- Benefits for institutions and processes—increased citizen engagement in public action, greater citizen ownership of public policies, improved public policies, improved ability of the executive to act.
- AI application example—AI solution providing an overview and synthesis based on debates and deliberations within a regional assembly.

*2.2.5 Participation in Public Policy Evaluation*

- Description—enabling all citizens to participate in the evaluation of public policies.
- Benefits for institutions and processes—stronger citizen engagement in public action, greater citizen ownership of public policies, improved public policies.
- AI use example—an AI tool to enable citizens to contribute to the evaluation of a public policy.

## 2.3 National Legislative Process

*2.3.1 Understanding and Engaging with the Legislative Process*

- Description—enabling all citizens to understand all aspects of the legislative process (committee work, parliamentary reports, plenary debates, bills, amendments and adopted laws) and to interact with legislators.
- Benefits for institutions and processes—improved citizen engagement and confidence in the legislative process and the laws it produces, improved parliamentary debates and laws passed.
- AI use example—an AI solution offering an overview and synthesis based on all the public content of the National Assembly (Parliament).

*2.3.2 Participation in Drafting Bills and Legislative Proposals*

- Description—enabling all citizens to participate in shaping legislation (scope, design, draft). (9).
- Benefits for institutions and processes—improved citizen engagement and trust in the law, better bills, strengthened capacity to pass effective laws.
- AI use example—AI platform to enable citizens to contribute to identifying key issues of a legislative bill.

*2.3.3 Understanding the Law*

- Description—helping all citizens better understand the law, its intricacies, and its implications.
- Benefits for institutions and processes—boosting citizen engagement and trust in the law.
- AI use example—AI tool providing an overview and synthesis of the key issues addressed in a piece of legislation.

### 2.4 Participatory Processes

*2.4.1 Understanding and Participating in Citizen Assemblies*

- Description—enabling all citizens to understand how a citizens' assembly works (agenda, expert hearings, debates, and recommendations) and to interact with it.
- Benefits for institutions and processes—greater citizen ownership of citizens' assembly recommendations, improved citizens' assembly recommendations, bolstered popular and political legitimacy of citizens' assembly recommendations, greater impact of citizens' assembly recommendations on legislative and executive processes.
- Example of AI application—AI solution to enable citizens to contribute to setting the agenda for a citizens' assembly.

*2.4.2 Participatory Budget*

- Description—empowering all citizens to propose and select projects for funding.
- Benefits for institutions and processes: increased transparency in local and national finances, boosting tax compliance and fostering innovative public spending.
- AI use example—an AI tool enabling co-creation of projects and aiding their comprehension to facilitate better-informed selections. (10)

## 3. Democratic Principles for LLM Supporting Participatory Civic Platforms

These new uses, some of which have great potential, should adhere to democratic principles. The LLMs that power the platforms should do the same. We will therefore consider an initial difficulty that is well known to philosophers: important concepts are contested. It is not that they are not accepted, but that they are subject to interpretative disputes. This is well known in moral and political philosophy.

We will then see how long-term projects have developed indicators that are able to assess democracy in its contexts of application. Thirdly, we will look at the limitations of this assessment project from our perspective and make new proposals for democratic principles for different forms of participation thanks to civic platforms, supported by different LLMs.

### 3.1 Democracy: "Contested" Concept and Comparability

One of the first difficulties is that democracy is a "contested" concept,[14] like most normative concepts. This is true not only at the empirical level, since not all citizens share the same definitions of democracy, but also at the level of theory or political philosophy. Moreover, we have already analyzed the attractiveness of different aspects of democratic legitimacy in a quantitative survey.[15]

However, this does not mean that we are bound to relativism. Indeed, debates on the definition and legitimacy of democracy are indebted to traditions of philosophical thought,[16] which are more or less attached to a particular set of values that they favor or even to certain components that make democratic life possible and desirable. We can go even further than agreeing on essential democratic principles and go so far as to define and translate them into dimensions and indicators for assessing the democratic quality of states around the world.

This is how the ambitious international comparative project V-DEM (Varieties of Democracy) was able to treat and classify, over two centuries, all the countries in the world from the angle of five principles, discussed in their book *Varieties of Democracy*.[17] V-DEM involved more than three thousand researchers on five continents and produced what is recognized as the most robust academic approach. They also carried out a state-of-the-art review of democratic principles (chapter 2, "Conceptual Scheme"). (11) Their list of seven principles (of which they retained only five for empirical analysis) borrows four from Robert Dahl:[18] (1) the electoral principle, with Dahl's concept of "polyarchy" in six distinctions: elected officials; free, fair, and frequent elections; and, to ensure that elections are free and fair, freedom of expression, access to alternative sources of information; associational autonomy; and inclusive citizenship; (2) the liberal principle,[19] concerned with protecting the rights of individuals against the oppression of an unrestricted majority; (3) the majoritarian principle, believing that the general popular will must be sovereign[20];and (4) the consensual principle.

Inspired by these theoretical proposals, the V-DEM authors defined their principles. To help interpret them, they proposed a question, core values, and attributes for each one.[21]

I. Electoral
   Question—Are important government offices filled by free and fair multiparty elections before a broad electorate?
   Core Values—Contestation/competition.
   Attributes—Inclusive suffrage, clean elections, elected officials, freedom of association, freedom of expression, and alternative information.
II. Liberal
   Question—Is power constrained and are individual rights guaranteed?

---

14. Walter Bryce Gallie, "Essentially Contested Concepts," *Proceedings of the Aristotelian Society* 56. David Collier, Fernando Daniel Hidalgo, and Andra Olivia Maciuceanu, "Essentially Contested Concepts: Debates and Applications," *Journal of Political Ideologies* 11, no. 3.
15. Reber, "Critical Citizenship and Democratic Legitimacy," *Philosophy and Social Criticism* 48, no. 9.
16. Frank Cunningham, *Theories of Democracy: A Critical Introduction*.
17. Michael Coppedge, John Gerring, Adam Glynn, et al., *Varieties of Democracy: Measuring Two Centuries of Political Change*.
18. Robert A. Dahl, *Polyarchy: Participation and Opposition*; *Democracy and Its Critics*; *On Democracy*; and *On Political Equality*.
19. David Held, *Models of Democracy*. Giovanni Sartori, *The Theory of Democracy Revisited*.
20. Walter Bagehot, *The English Constitution*.
21. Coppedge, et al., *Varieties of Democracy*.

Core Values—Individual liberty, checks and balances, constitutionalism.
Attributes—Civil liberties, judicial independence, legislative independence.

III. Majoritarian
Question—Does the majority rule via one party, and does it dominate policy making?
Core Values—Majority rule, power concentration, efficient decision-making, responsible party government.
Attributes—Power-concentrating institutions, power-centralizing institutions, simple majority decision-making.

IV. Consensual
Question—Do numerous, independent, and diverse groups and institutions participate in policy making?
Core Values— (12) Voice and representation of all groups, power dispersion, power sharing.
Attributes—Power-dispersing institutions, power-decentralizing institutions, supermajority decision-making.

V. Participatory
Question—Do citizens participate in political decision-making?
Core Values—Direct, active participation in decision-making by the people.
Attributes—High turnout, mechanisms of direct democracy, civil society activism, local democracy.

VI. Deliberative
Question—Are political decisions the product of public deliberation based on reasoned and rational justification?
Core Values—Reasoned debate and rational arguments, consultation.
Attributes—Public debate, respectful, open-minded discussions, reasoned justification with reference to the public good, consultative institutions.

VII. Egalitarian
Question—Are all citizens equally capable to use their political rights?
Core Values—Equal political capabilities.
Attributes—Equal protection of rights and freedoms, equal distribution of politically relevant resources, equal access to power.

Given the DC project's perspective, we have first reconfigured these seven principles and established priorities in light of the new possibilities enabled by AI. Indeed, we have selected democratic uses associated with new possibilities for participation thanks to digital networks. Furthermore, since AI is associated with them, we are interested in knowledge-based possibilities. These possibilities are very different, based on statistics and calculations.

However, AI or knowledge-based possibilities appear above all as "literary technologies" and discourse order.[22] AI is called upon to translate, summarize, and synthesize. It is even there to help with comprehension. We speak hastily of conversations thanks to conversational robots based on questions asked by humans. Conversation is richer and notoriously political.[23] In view of all these mobilized capacities, it is the principle of deliberation that is the most relevant in V-DEM's list of principles.

Secondly, these principles are not of the same order. Guaranteeing freedoms, a liberal principle, is a prerequisite for the exercise of democracy but is not specific to it or at the heart of its exercise. The electoral principle is included as a phase in (13) our project. The same applies to the egalitarian principle, which needs to be qualified. In our project, this equality translates into equal access to the thirteen uses of democracy.

---

22. Reber, *Responsible Deliberation*.
23. Reber, *Responsible Deliberation*. Michael Schudson, "Why Conversation Is Not the Soul of Democracy," *Critical Studies in Mass Communication* 14, no. 4.

With new forms of participation for citizens, instrumented and supervised by moderators and facilitators, these principles apply differently depending on the agents involved and the uses considered, as we shall see. A platform moderator will have to be neutral towards contributions, while participants will be able to put forward their point of view in a way that respects ethical and political pluralism. It may be possible for the moderator to be charged with ensuring that a minority position is heard and that the moderator does take action differentially based upon a position (or maybe the position's relation to the space of possible positions and those actually taken within a context).

### 3.2 Proposal of a List of Five Democratic Principles

Of course, we are not about to reinvent the wheel. However, while V-DEM's list is interesting, these principles do not belong on the same level. The two principles closest to our concerns are participatory and deliberative. They also potentially modify the electoral, legislative, and executive uses of the electoral principle. To be sure, AI associated with digital networks (social networks/social networking sites/digital platforms/social media platforms) directly affects the freedoms that liberal principles are supposed to guarantee, and we need to watch out for this destabilization and the shameless plundering of users' private data often unbeknownst to the users themselves.[24] However, these principles are only prerequisites for the democratic exercise and its difficulty.

The deliberative principle is minimal in V-DEM. As far as indicators are concerned, V-DEM simply adopts the "Deliberative quality index."[25] This index has many limitations, which are acknowledged by the authors themselves. If we are optimistic, we can hope that AI will help instrument, operationalize, or objectify deliberation. For the time being, they are far from being able to do so. (14)

Above all, the V-DEM list overlooks the need for pluralism, which lies at the heart of the problem of political justice.[26] The means of taking into account[27] ethical pluralism (e.g., for conceptions of justice) and epistemic pluralism (for establishing facts or probabilities) involves all kinds of communicative capacities that AIs can implement.

---

24. Pauline Elie. "Du nomos des identités au dominium de la personne. Analyser l'identité en droit: Protéger un nouveau territoire à l'ère numérique." Zuboff, *Age of Surveillance Capitalism*.
25. Marco R. Steenbergen, André Bächtiger, Markus Spörndli, and Jürg Steiner, "Measuring Deliberation: A Discourse Quality Index," *Comparative European Politics* 1. Steiner, *The Foundations of Deliberative Democracy: Empirical Research and Normative Implications*. Steiner, Bächtiger, Spörndli, and Steenbergen, *Deliberative Politics in Action: Analysing Parliamentary Discourse*. Steiner, Maria Clara Jaramillo, Rousiley C. M. Maia, and Simona Mameli, *Deliberation Across Deep Divisions: Transformative Moments*. Bächtiger, John S. Dryzek, Jane Mansbridge, and Mark E. Warren, eds., *The Oxford Handbook of Deliberative Democracy.*
26. John Rawls, *A Theory of Justice*; "Political Liberalism: Reply to Habermas," *Journal of Philosophy* 92, no. 3; *Justice as Fairness: A Restatement*; and *Political Liberalism*. Jürgen Habermas, *The Theory of Communicative Action, Volume 1: Reason and the Rationalization of Society*; *The Theory of Communicative Action, Volume 2: Lifeworld and System: A Critique of Functionalist Reason*; "Reconciliation Through the Public Use of Reason: Remarks on John Rawls's Political Liberalism," *Journal of Philosophy* 92, no. 3; *Between Facts and Norms: Contributions to a Discourse Theory of Law and Democracy*; and *The Inclusion of the Other: Studies in Political Theory*.
27. Reber, *Responsible Deliberation*.

The majoritarian and consensual principles are also interesting, as they are components of the discussion and deliberation required to reach or agree on decisions. However, the V-DEM project does not use these two principles. We mustn't be satisfied with consensus as the only mode of agreement; there are others, such as conflict clarification.[28] Nor should we forget disagreements, which can be of high quality, necessary passages, or even conditions calling precisely for deliberation. Moreover, deliberation must take into account the possibility of criticism, which is, in a way, its driving force. This criticism is more sophisticated depending on whether we react, evaluate, or make an alternative proposal, while respecting the pluralism that guarantees the existence of critics.[29]

There is another omission from the V-DEM list. The granting of greater or lesser opportunities for participation implies *responsibilities* in proportion to the powers granted. This means shared responsibilities (e.g., responding to a question or objection in a public debate between citizens) and accountability (e.g., as a member of an executive upon receipt of a report from conventioneers). We therefore exercise common but differentiated responsibilities. The list of priority democratic principles given the DC project could therefore be:

1. Participation.
2. Political and ethical pluralisms (in a sense that goes beyond plurality).
3. Responsibilities (e.g., capacity and accountability).
4. Deliberation.
5. Agreements/Disagreements

### 3.3 Democratic Principles and Indicators

(15) These principles can be broken down into indicators, which are listed below. I propose nineteen of them.

*3.3.1 Participation.*
Here we answer the question: "Who participates?" The main focus is on the citizens who are invited and at the center of the debates. However, the organizers and suppliers of tools shape these debates. They too must ensure that these principles are respected, through a set of complementary principles. Some organizers, the moderators for example, directly ensure that the principles guaranteeing the democratic quality of the debates are applied. The indicators are:

> *(Participation 1.1.)* Free access—This freedom must nevertheless take into account a charter favoring the serenity and security of democratic exchanges, and respect for other participants.
> *(Participation 1.2.)* Equal participation and consideration—Everyone must be able to make their position heard and defended, in the language of their choice, whether directly or by consent. Indeed, this is more important than just equal speaking time, where not everyone needs the same amount of time to express themselves.
> *(Participation 1.3.)* Descriptive diversity (plurality)—This may be dependent on socio-demographic criteria (e.g., those of the French citizens' conventions: age, gender, profession, level of education, type and density of housing, but also partisan affinities, or

28. Amy Gutmann and Dennis Thompson, *Democracy and Disagreement: Why Moral Conflict Cannot Be Avoided in Politics, and What Should Be Done About It*.
29. Reber, "Critical Citizenship."

attitudes to the subjects dealt with, or assignment according to the subjects dealt with and the decisions that may be taken).

*3.3.2 Political and Ethical Pluralisms*

Here we answer the question of the recognition of reasonable pluralism as a fact (Rawls) but also the question of the pluralism of moral theories, because moral evaluations and disagreements are dependent on the existence of several types of ethical justification.[30] It is neither relativism (which says that moral and political problems are intractable as such), nor monism (which believes that only one moral or political position is possible). While this pluralism is guaranteed by law and institutions (Habermas), it is often not accepted by many citizens, whether responsible or not. The indicators are:

*(Political and Ethical Pluralisms 2.1.)* Is political pluralism recognized?
*(Political and Ethical Pluralisms 2.2.)* Is ethical pluralism recognized? (16)
*(Political and Ethical Pluralisms 2.3.)* Are all ethical and/or political positions identified?

*3.3.3 Responsibilities*

According to the powers devolved by the democratic usage in question:

*(Responsibilities 3.1.)* Do participants have the capacities and the possibility of increasing them to be able to contribute to the expected tasks?
*(Responsibilities 3.2.)* Do participants sincerely commit themselves (as if they had to submit to it themselves) to the proposals they defend, taking into account their consequences?
*(Responsibilities 3.3.)* Are the participants, or the group they form (e.g., a convention), accountable to the authorities and audiences concerned?
*(Responsibilities 3.4.)* Are exchanges *responsive*, with modalities specific to different uses, both internally and externally (different audiences)?

*3.3.4 Deliberation*

This principle concerns the quality of democratic deliberation.

*(Deliberation 4.1.)* What is the number of exchanges?
*(Deliberation 4.2.)* Are they civil and respectful of people?
*(Deliberation 4.3.)* Are claims given justified?
*(Deliberation 4.4.)* Are positions changing: individually? collectively?
*(Deliberation 4.5.)* Is there room for both individual and collective deliberation?
*(Deliberation 4.6.)* Is there room for both ethical and political deliberation?

*3.3.5Agreements/Disagreements*

*(Agreements/Disagreements 5.1.)* For each use, are agreements—or disagreements—foreseen?
*(Agreements/Disagreements 5.2.)* Are these agreements—or disagreements—recognized as legitimate by the participants?
*(Agreements/Disagreements 5.3.)* Are democratic means provided to achieve them?

---

30. Shelly Kagan, *Normative Ethics*. John Kekes, *The Morality of Pluralism*. Reber, "Technology Assessment as Policy Analysis: From Expert Advice to Participatory Approaches," in *Handbook of Public Policy Analysis: Theory, Politics and Methods*; "Governance between Precaution and Pluralism," *International Social Science Journal* 211 and 212; and *Precautionary Principle, Pluralism, Deliberation: Science and Ethics*.

We can then list the means provided: voting (differently qualified), deliberation, for agreements (consensus, compromise, acquiescence to difference) or different disagreements (deliberative disagreements). These methods can be combined. For example, deliberate then vote, or deliberate, vote, and deliberate again. They can also be chained, from disagreement to agreement. In order to help to interpret these (17) indicators, sub-questions can be identified to enable evaluation using more detailed standards.

### 3.4. Democratic Indicators and AI-Assisted Roles

We will interpret these indicators for five possible roles of AI:

- Moderation—an AI decides if a citizen contribution is acceptable or not.
- Translation—an AI enables a multilingual debate with automatic translation.
- Summarization—an AI summarize numerous citizen contributions into a synthetical summary.
- Writing assistant (formulation and reformulation)—an AI helps a citizen to improve the formulation of her or his contribution. An AI suggests a reformulation of a citizen contribution before it is shared with other citizens to make it more acceptable by other citizens.
- Facilitation: an AI ensures that a debate among citizens is civil and constructive.

For instance, if we take the *Participation 1.2* indicator on equal consideration, for the role of moderation AI-assisted we will have to check that:

a. Moderation:
   i. Is the AI discriminatory towards certain citizen groups?
   ii. Is the AI more severe on certain ideas?
b. Translation:
   i. Is the AI translating better or worse certain citizen groups?
   ii. Is the AI altering the translation of certain ideas?
   iii. Is the AI translating better certain languages?
c. Summarization:
   i. Is the AI giving more or less space in the summary to certain ideas?
   ii. Is the AI giving more or less space in the summary to certain citizen groups
d. Writing Assistant (formulation and reformulation):
   i. Is the AI altering more certain ideas than others?
   ii. Is the AI reformulating certain citizen groups?
e. Facilitation:
   i. Is the AI favoring or intervening more on certain ideas?
   ii. Is the AI using its facilitating role more to favor certain citizen groups?

This analysis makes three potential biases emerge: (18)

- representation biases—discriminating against certain citizen groups on age, gender, ethnicity, or social categories.
- political biases—discriminating or favoring certain ideas, positions, or ideologies.
- language biases—discriminating for or against certain languages.

This example demonstrates the methodology we propose to apply to bridge the gap between the theoretical principles, indicators, and the practical application and assessment of AI for democracy.

## Conclusion

AI, and more specifically LLMs, must be subject to democratic principles, which make it possible, on the one hand, to evaluate the models themselves as well as their current uses, and, on the other, to propose new resources so that LLMs can serve democracy or even help to realize some of its promises. The Democratic Commons project has therefore identified five principles:

1. Participation
2. Political and ethical pluralisms
3. Responsibilities (e.g. capacity and accountability)
4. Deliberation
5. Agreements/Disagreements

It has extracted 19 indicators that are associated with questions to help translate them into standards according to the disciplinary practices and objects studied. In fact, these interpretations are indebted to the scientific contexts and objects under consideration.

In order to help to interpret these indicators, sub-questions have been identified to enable evaluation using more detailed standards for five roles of AI: moderation, translation, summarization, writing assistant (formulation and reformulation), and facilitation.

In the course of the project, these indicators will be interpreted to assess thirteen different types of democratic uses in four phases of political life: elections, the exercise of power, the legislative process, and participation in mini-publics.

The variety of participatory democratic uses envisaged implies different modes of application for the principles and their indicators. Indeed, different uses combine different modes of communication. Expressing an idea to inspire an electoral program is not the same as co-constructing a public policy or, even less, participating in a convention. Each usage designates different cognitive and social activities: expression, inspiration, prioritization, understanding (of institutional functioning, institutional design, decision-making processes, drafting of legislation, public policy proposals), (19) comparison, monitoring (reading debates), evaluation (of program execution, public policy), interaction, drafting (for example, of proposals as part of a citizens' convention), participation in all the activities of a citizens' convention, budget proposal, or selection of budget projects. These activities induce distinct critical capacities of participants: reactive, evaluative, propositional, and capable of recognizing political pluralism (the coexistence of critical possibilities.[31]

Similarly, uses are sometimes limited to a single *expression*. At other times, they give rise to *interaction*. They are arranged between two poles, ranging from the most minimal, the survey, to the most maximal: the citizens' convention.

Finally, the powers they confer call for commensurate responsibilities in terms of *accountability*[32] for the proposals and decisions made. They also require different *capacities*. A final sense of responsibility,[33] such as *responsiveness*, is essential in the event of interaction, so that subsequent communications take into account previous responses by others.

This is one of the weaknesses of the asynchronous nature of online participative experiences. In addition, the question arises of lifting anonymity if significant decision-making power is devolved.

---

31. Reber, "Critical Citizenship."
32. Mark Bovens, Robert E. Goodin, and Thomas Schillemans, eds. *The Oxford Handbook of Public Accountability.*
33. Sophie Pellé and Reber, *From Research Ethics to Responsible Innovation and Research*, and "Responsible Innovation in the Light of Moral Responsibility," *Journal on Chain and Network Science* 15, no. 2. Reber, "RRI as Inheritor of Deliberative Democracy and the Precautionary Principle," *Journal of Responsible Innovation*" 5, no. 1.

In terms of research, the DC project could make a major contribution to empirical, normative, and educational work on the most elaborate form of democracy, that of deliberative democracy. Indeed, work on the deliberative system[34] remains very vague[35] at the risk of making deliberation lose all its distinctive features.[36] Here we would have the possibility of putting together different forms of participation in parts of democratic systems to be made more deliberative. We could then hope that the democratic system as a whole would benefit. More ambitiously, we could also hope to capture a democratic system as a whole through the accumulation of these thirteen democratic uses. We would then have provided the means to improve what we call a deliberative system. (20).

---

34. John Parkinson and Jane Mansbridge, eds. *Deliberative Systems: Deliberative Democracy at the Large Scale*. John Parkinson, "Deliberative Systems," in *The Oxford Handbook of Deliberative Democracy*."
35. David Owen and Graham Smith, "Survey Article: Deliberation, Democracy, and the Systemic Turn," *Journal of Political Philosophy* 23, no. 2.
36. Robert E. Goodin, "If Deliberation Is Everything, Maybe It's Nothing," in *The Oxford Handbook of Deliberative Democracy*.